\documentclass[runningheads]{llncs}

\usepackage{amsmath}
\usepackage[T1]{fontenc}
\usepackage{graphicx}
\usepackage{enumitem}
\usepackage{xcolor}   
\usepackage{xspace}
\usepackage[indent]{parskip}
\usepackage{booktabs}
\usepackage{colortbl}
\usepackage[pagebackref,breaklinks,colorlinks]{hyperref}
\usepackage{cleveref}

\def\modelname{HLSS\xspace}
\definecolor{Gray}{gray}{0.90}

\DeclareMathOperator*{\argmax}{arg\,max}

\newcommand{\skipsum}[1]{%
  \mathop{{\vphantom{\sum}}_{#1}\kern-\scriptspace}\!\sum\nolimits
}

\begin{document}

\title{Hierarchical Text-to-Vision Self Supervised Alignment for Improved Histopathology Representation Learning}
\titlerunning{HLSS: Hierarchical Language-tied Self Supervision}

\author{Hasindri Watawana\inst{1}\and
Kanchana Ranasinghe\inst{2}\and
Tariq Mahmood\inst{5}\and
Muzammal Naseer\inst{1}\and
Salman Khan\inst{1,4}\and
Fahad Shahbaz Khan\inst{1,3}
}

\authorrunning{H. Watawana et al.}

\institute{Mohamed Bin Zayed University of Artificial Intelligence \and
Stony Brook University \and
Link$\ddot{o}$ping University \and
Australian National University \and
Shaukat Khanum Cancer Hospital, Pakistan}

\maketitle              

\begin{abstract}

Self-supervised representation learning has been highly promising 
for histopathology image analysis with numerous approaches leveraging their patient-slide-patch hierarchy to learn better representations. In this paper, we explore how the combination of domain specific natural language information with such hierarchical visual representations can benefit rich representation learning for medical image tasks. Building on automated language description generation for features visible in histopathology images, we present a novel language-tied self-supervised learning framework, Hierarchical Language-tied Self-Supervision (\modelname) for histopathology images. We explore contrastive objectives and granular language description based text alignment at multiple hierarchies to inject language modality information into the visual representations. Our resulting model achieves state-of-the-art performance on two medical imaging benchmarks, OpenSRH and TCGA datasets. Our framework also provides better interpretability with our language aligned representation space. Code is available at \href{https://github.com/Hasindri/HLSS}{\textcolor{blue!50!cyan}{https://github.com/Hasindri/HLSS}}.


\end{abstract}

\section{Introduction}

Self-supervised learning (SSL) has showcased remarkable outcomes for vision tasks \cite{he2022masked,caron2021emerging}, with recent extensions to medical imaging tasks proving highly successful \cite{Jiang2023HierarchicalDL}, especially given the expensive and difficult nature of medical image annotation due to necessity of domain-specific expertise knowledge. 

In constrast to natural images, medical images contain unique imaging patterns. In clinical studies, it is common to sample multiple gigapixel range image \textit{slides} from a single \textit{patient}, followed by analysis of smaller sub regions of slides, referred as \textit{patches}. This creates a patient, slide, patch hierarchy in captured data where all samples from a single patient correspond to a common diagnosis. Multiple existing works leverage this hierarchy in the visual modality to learn self-supervised representations of histopathology images \cite{Jiang2023HierarchicalDL,chen2022scaling}.

With the recent advancement of vision-text alignment research and highly transferable vision language models (VLMs) \cite{Radford2021LearningTV,li2022blip}, multiple histopathology SSL works go beyond raw visual information, leveraging natural language to learn more generic representations, zero-shot capabilities, and improved interpretability \cite{lu2023visual,wang2022medclip,lu2023towards}. However, utilising hierarchy in terms of language has not yet been done in any recent vision-language SSL works in histopathology, especially given most of these works are built upon image-text paired datasets manually annotated or automatically captioned at only patch level \cite{ikezogwo2024quilt,gamper2021multiple}. 

In our work, we bridge this gap by exploring hierarchy in both vision and language modalities. We propose a novel framework for hierarchical text-to-vision alignment for language guided visual representation learning named \textbf{\modelname (Hierarchical Language-tied Self Supervision)} which extends the self supervised learning objectives across three levels of hierarchy: patient, slide and patch.
In contrast to existing histopathology VLM approaches that require a sample specific description per each image, we are the first to use a fixed set of textual descriptions depicting dataset specific characteristics, for language guidance. First, we utilise pre-trained LLMs \cite{brown2020language} containing extensive world knowledge to extract a fixed set of visual characteristics flagged as useful for a diagnosis in a given dataset type, for each level of the hierarchy. Then a textual description is generated per each attribute. Since the three sets of descriptions describe the images at three different granularities; patch level describes more fine grained features while patient level describes features related to overall diagnosis; we refer to them as granular language descriptions. This entire process is automated and is followed by verification from a human expert in histopathology.
Collected textual descriptions are encoded using a CLIP text encoder \cite{Radford2021LearningTV} and the resulting text vectors are used to construct a hierarchical text-to-vision alignment objective which is combined with a hierarchical vision contrastive objective. The resulting framework, which we named \modelname, learns representations achieving state-of-the-art performance on downstream medical image classification tasks.

\noindent We summarize our key contributions as follows:
\begin{enumerate}[leftmargin=2.5em,noitemsep,topsep=0.3em,itemsep=-1.0ex,partopsep=0ex,parsep=1ex]
    \item Automated Generation of dataset specific granular \textit{characteristic-description} text pairs that can describe histopathology images at a multitude of levels
    \item Hierarchical text-to-vision alignment for self-supervised representation learning on histopathology images
    \item A language guided framework that utilise dataset specific characteristic descriptions instead of sample specific captions
\end{enumerate} 
Evaluations on two downstream histopathology benchmarks, OpenSRH and TCGA, demonstrate state-of-the-art performance of our \modelname framework. 

\section{Related Work}

\noindent\textbf{Self Supervised Learning in Histopathology:}
Histopathology is the study of diseases by examining tissue samples obtained from patients. In clinical medicine and biomedical research, histopathology studies are strongly supported by rapidly advancing microscopy techniques (e.g. light microscopy, stimulated raman microscopy), used to image the tissue samples. Digitization of these imaging processes have resulted in bulks of digital image patient data, mostly left unannotated due to the requirement of specific expertise. Therefore, deep learning methods in histopathology have adopted self-supervised pretraining as an essential pre-conditioner for training visual recognition models. Contrastive learning \cite{ciga2022self}, reconstruction from partial signals \cite{boyd2021self}, DINO-based knowledge distillation \cite{chen2022self} are some of the SSL work in histopathology.

\noindent\textbf{Hierarchical Self Supervised Learning in Histopathology:} 
Multiple recent works leverage the inherent hierarchy in histopathology data to learn useful representations. A hierarchical image contrastive objective to learn patch representations is presented in \cite{Jiang2023HierarchicalDL}. A weakly supervised hierarchical pyramid vision transformer (ViT) for slide representation learning starting from patch level tokens is explored in \cite{chen2022scaling}. In contrast to these approaches utilising hierarchy only within vision modality, our proposed \modelname is extended to language.

\noindent\textbf{Vision Language Models in Histopathology:}
Numerous recent works in histopathology combine language with self-supervised objectives, mostly basing off image-text contrastive objectives applied on paired image-caption data \cite{ikezogwo2024quilt,lu2023towards,lu2023visual,wang2022medclip}. However this demands sample-specific textual captions that are expensive for the medical domain. At the same time, these methods do not explore the hierarchical relations inherent to the data. To the best of our knowledge, our work is a first to introduce a hierarchical vision-language representation learning approach. Moreover, our setup utilizes dataset-specific characteristic descriptions, avoiding the need for expensive per image textual captions for training.

\section{Methodology}
In this section, we present our approach, \modelname, that learns language-guided self-supervised representations for histopathology images that are also interpretable for clinical decision-making. 
We leverage the inherent hierarchical structure of histopathology data, across both vision and language modalities, and construct two novel language-tied self-supervision objectives, Hierarchical Vision Contrastive (HVC) Loss and Hierarchical Text-to-Vision Alignment (HA) Loss. Additionally, we propose two architectural components, Positive Pairing Module (PPM) and Cross-Modal Alignment Module (CAM) for efficient implementation of our objectives. In the following sections, we discuss some key characteristics of histopathology data, layout the architecture of our framework, introduce our two proposed learning objectives in detail, and describe our overall framework. To the best of our knowledge, in the histopathology imaging domain, we are the first to connect natural language with hierarchy aware self-supervised learning.

\subsection{Background}
\label{subsec:background}

Histopathology domain vision tasks generally involve hierarchical data extraction \cite{Jiang2023HierarchicalDL}, resulting in an inherent hierarchy for the visual information. A single Whole Slide Image (WSI), referred as a \textit{slide}, could span gigapixel scales, motivating most computer vision approaches to use sub-sampled fields-of-view (e.g., 256×256 pixels region), referred as \textit{patches}. These patches could belong to a single slide or different slides from the same patient. Overall, this results in an inherent three-tier data hierarchy. Interestingly, each level contains some unique visual characteristics which can also be described sufficiently using natural language, for example, 
\begin{enumerate}
[leftmargin=2.5em,noitemsep,topsep=0.3em,itemsep=-1.0ex,partopsep=0ex,parsep=1ex]
    \item \textbf{Patch:} cellular features such as cellularity, DNA localization, cell density.
    \item \textbf{Slide:} tissue features such as tissue architecture, cell infiltration.
    \item \textbf{Patient:} diagnosis features such as tumor heterogeneity, lipid distribution.
\end{enumerate} 
This provides the two-fold motivation for our proposal: leveraging hierarchy and connecting natural language to learn stronger representations. 

\subsection{Architecture}
\label{subsec:arch}
\begin{figure}[t]
    \centering
    \includegraphics[width=0.95\linewidth]{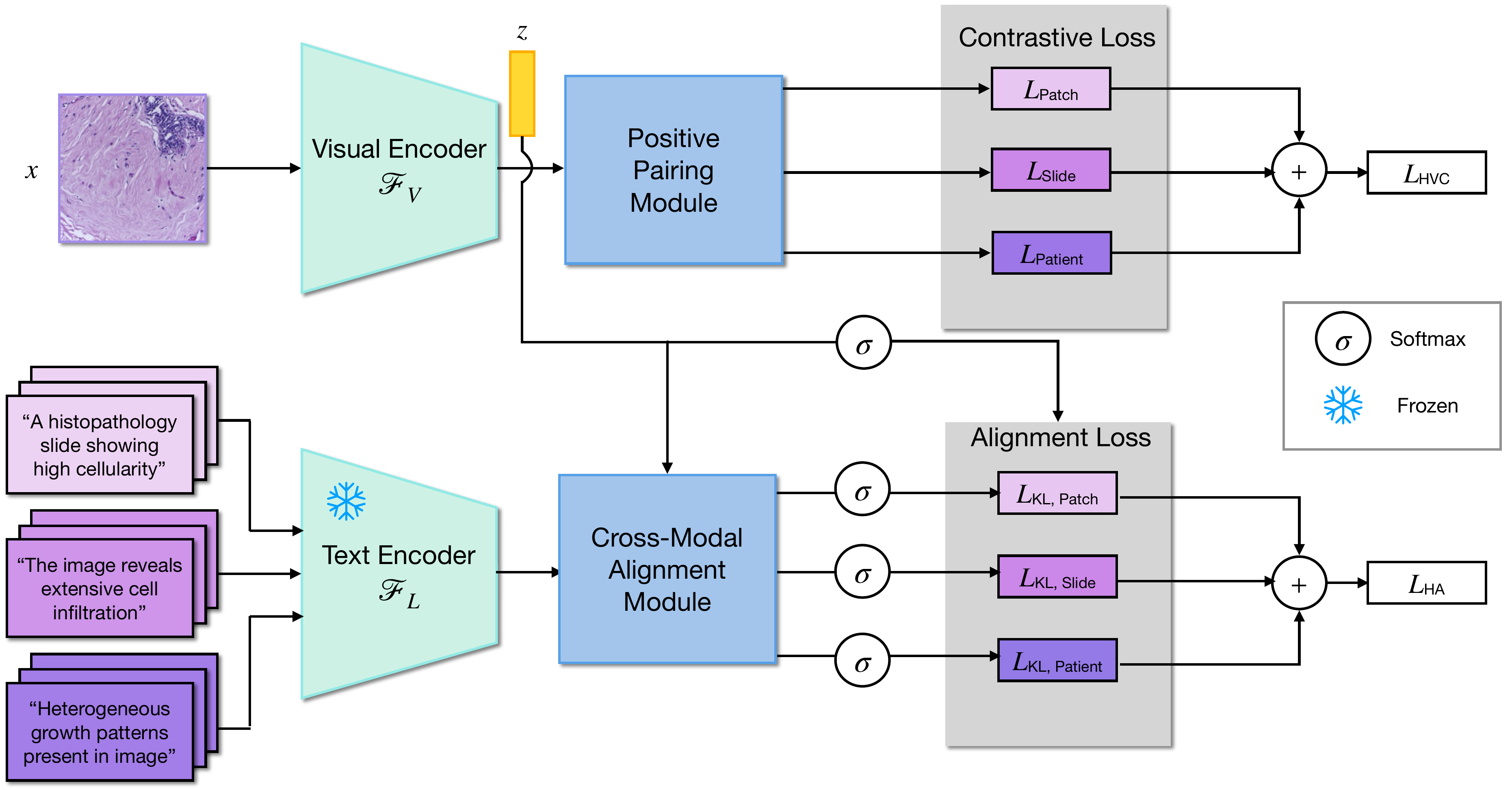}
    \vspace{-1.0em}
    \caption{\textbf{Overview of \modelname Architecture}}
    \label{fig:arch}
    \vspace{-1.5em}
\end{figure}

Our \modelname processes patches, $x \in \mathbf{R}^{(H,W,C)}$ where $H=W=224$ and $C=3$, to output representations, $z \in \mathbf{R}^D$ where $D=1024$, which are used in downstream tasks. The inherent data hierarchy allows sampling patches belonging to individual levels, as illustrated in \cref{app:hist_imaging}. 
Therein, our setup processes $n_{s}*n_{p}*n_{a}$ patches at each iteration, where $n_{s}$ slides belong to a common patient, $n_{p}$ patches are sampled from each slide, and data augmentations provide $n_{a}$ views per patch. A view here refers to a visually augmented version of a patch. 
We use a ResNet-50 \cite{He2015DeepRL} as our visual encoder with CLIP pretraining \cite{Radford2021LearningTV}, $\mathcal{F}_V$ to obtain $z \in \mathbf{R}^D$. 
Our proposed positive pairing module (PPM) projects $z$ to language guided representation spaces motivated from \cite{Ranasinghe2023LanguagebasedAC} obtaining features specific to each level as $z'_{\text{patient}}, z'_{\text{slide}}, z'_{\text{patch}}$. These features are processed by our proposed hierarchical vision contrastive (HVC) objective to provide the first learning signal. 
Additionally, granular textual descriptions relevant to each hierarchy are processed through a language encoder from CLIP \cite{Radford2021LearningTV}, $\mathcal{F}_L$ to obtain language modality features, $t_{\text{patient}}, t_{\text{slide}}, t_{\text{patch}}$. Our cross-modal alignment (CAM) module outputs modified features  $t'_{\text{patient}}, t'_{\text{slide}}, t'_{\text{patch}}$ which processed by our Hierarchical Text-to-Vision Alignment (HA) objective, providing a second learning signal. 
We illustrate this overall architecture in \Cref{fig:arch}. 

\subsection{Hierarchical Vision Contrastive Objective}
\label{subsec:contrast}
Self-supervised contrastive objectives commonly operate on features that are projected to a secondary feature space, which is considered to learn more generic representations \cite{Balestriero2023ACO}. Given the hierarchy of our data, we propose distinct secondary feature spaces for each level. Additionally, motivated by \cite{Ranasinghe2023LanguagebasedAC}, we propose language-guided construction of these feature spaces. This combined setup is implemented in our proposed Positive Pairing Module (PPM). The resulting hierarchical representations are processed separately using our hierarchical vision contrastive (HVC) loss to provide a suitable learning signal.

\noindent \textbf{Positive Pairing Module:}
We first separate visual encoder outputs, $z$, to individual levels and process each through level-specific projection layers. These projection layers are implemented as linear layers and output $z'_{\text{patient}}, z'_{\text{slide}}, z'_{\text{patch}}$ where each $z' \in \mathbf{R}^{(m, 128)}$ with $m$ equal to ($n_s\cdot n_p \cdot n_a$), $ (n_p \cdot n_a$), and $n_a$ respectively. The projection layers are initialized with textual vectors (extracted from $\mathcal{F}_L$) such that each axis of the 128-dimensional feature space corresponds to some textual description of a histopathology characteristic (details in \cref{subsec:align}).  

\noindent \textbf{Loss Definition:}
Note how each $z'$ above corresponds to more than one patch. Let us refer to features of a single patch as $z'_i$.
Given these projected $z'_i$, we consider $z'_i$ of common levels positives and others negatives, and we apply a contrastive objective from \cite{Khosla2020SupervisedCL}, 

\begin{equation}
\footnotesize
\label{eq:con_loss}
    L_\text{con}(z_i, Z_{pi}, Z) = \frac{1}{|Z_{pi}|} \sum_{z_k \in Z_{pi}} 
    \log \frac{\exp \left( z_i \cdot z_k \right / \tau)}
    {\sum_{z_j \in Z \backslash \{z_i\}} \exp\left( z_i \cdot z_j \right / \tau)}
\end{equation}
that supports multiple positives, where $Z_{pi}$ is the set of positives of $z_i$ (excluding $z_i$) and $Z$ is the set of all $z'$ respectively. Here each of $z'_{\text{patient}}, z'_{\text{slide}}, z'_{\text{patch}}$ corresponds to more than one patch (i.e. $n_s\cdot n_p \cdot n_a$, $ n_p \cdot n_a$, and $n_a$ patches respectively) while $z_i$ in \cref{eq:con_loss} refers to features of a single patch. Applying this loss at each hierarchical level we obtain three different objectives, 
{\footnotesize
\begin{align}
    L_\text{Patch} &= \skipsum{}_{z_i \in z'_\text{patch}}  L_\text{con}(z_i, z'_\text{patch} \backslash \{z_i\}, Z) \\
    L_\text{Slide} &= \skipsum{}_{z_i \in z'_\text{slide}}  L_\text{con}(z_i, z'_\text{slide} \backslash \{z_i\}, Z) \\
    L_\text{Patient} &= \skipsum{}_{z_i \in z'_\text{patient}}  L_\text{con}(z_i, z'_\text{patient} \backslash \{z_i\}, Z)
\end{align}
}
We combine these terms to define our first training objective HVC loss as, 
\begin{equation}
    L_\text{HVC} = L_\text{Patch} + L_\text{Slide} + L_\text{Patient}
\end{equation}

\begin{figure}[t]
    \centering
    \includegraphics[width=0.90\linewidth]{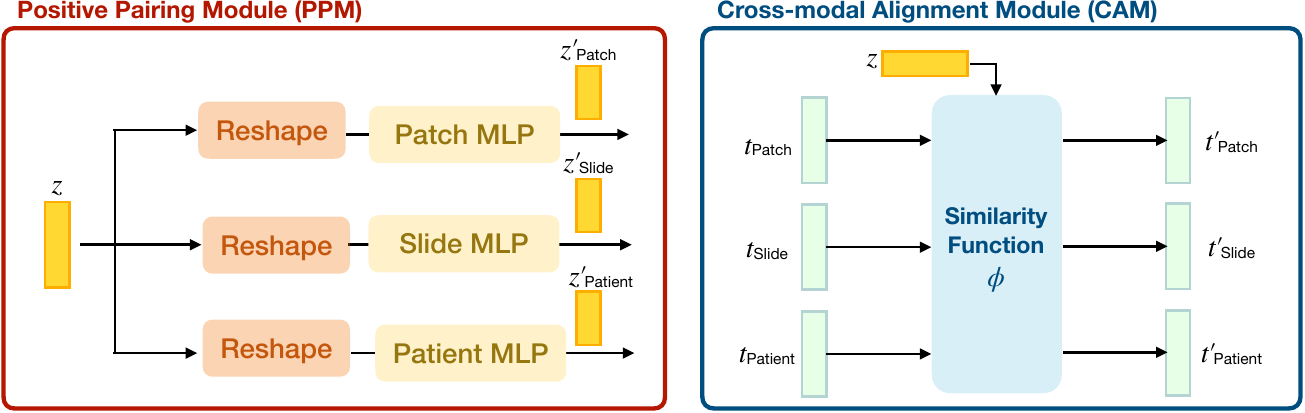}
    \vspace{-1.0em}
    \caption{
    We illustrate the operations within our proposed Positive Pairing Module (left) and Cross-Modal Alignment Module (right).  
    }
    \label{fig:CAMPPM}
    \vspace{-1.0em}
\end{figure}

\subsection{Hierarchical Text-to-Vision Alignment}
\label{subsec:align}

Our second training objective focuses on explicit alignment of cross-modal representations. We utilize a pretrained large language model (LLM) \cite{brown2020language} that contains both world knowledge and domain specific awareness to generate descriptions of visual characteristics corresponding to each level of our hierarchy. These generated descriptions are additionally verified by human experts (histopathology specialists) to ensure both their meaningfulness and interpretability when proposing diagnoses in downstream tasks. 

\noindent \textbf{Granular Language Descriptions:} 
We prompt ChatGPT in a multi-stage manner to generate a large set of visual descriptions which is reduced to 128 characteristics per level by eliminating redundancy and invalid responses manually (with human expert supervision). Each characteristic would be represented by 4 natural language descriptions. 

\noindent \textbf{Cross-Modal Alignment:} 
Given our language dataset of medical characteristic descriptions, we utilize our text encoder $\mathcal{F}_L$ to process these descriptions and extract language embeddings.The averaged embedding (across four descriptions) of each characteristic, $t_{\text{patient}}, t_{\text{slide}}, t_{\text{patch}}$ is calculated, where each $t \in \mathbf{R}^{(128,D)}$ with $D=1024$. For a given patch, using its visual embedding $z$ we select a best matching text embedding $t'_i = \argmax_{t_k \in t_i} \langle t_k, z \rangle$ where $\langle \cdot, \cdot \rangle$ is cosine-similarity and $t_k \in t_i$ indexes along the characteristic (128) dimension with $t_k \in \mathcal{R}^D$. 

\noindent \textbf{Hierarchical Alignment Objective:} 
We utilize the language embeddings $t'_i$ output from CAM to enforce distillation into our visual embeddings, $z$. Therein, our hierarchical alignment (HA) loss is defined using a KL-divergence loss,
\vspace{-0.5em}
\begin{align}
    L_\text{HA} = L_\text{KL}(z, t'_\text{patient}) + L_\text{KL}(z, t'_\text{slide}) + L_\text{KL}(z, t'_\text{patch})
\end{align} 
where $L_\text{KL}$ applies KL-divergence loss with suitable softmax normalization. 

\vspace{0.5em} \noindent
Therein, our overall self supervised learning objective can be summarised as,
\vspace{-0.5em}
\begin{equation}
L_\text{\modelname} = L_\text{HVC} + L_\text{HA}
\end{equation}

\section{Experiments}
\label{sec:exp}

\subsection{Experimental Setup}

\noindent\textbf{Datasets:}
We perform training (self-supervised) and evaluation on two benchmark histopathology image datasets: OpenSRH \cite{jiang2022opensrh} and TCGA.  
OpenSRH (Stimulated Raman Histology) is a public dataset of clinical SRH images of 300+ brain tumor patients with classes consisting of normal brain tissue and 6 different brain tumor diagnoses. We train on their train split containing 225k images and evaluate on the validation split containing 58k images. 
In TCGA brain cancer dataset, we utilise TCGA-LGG and TCGA-GBM subsets containing brain tumor samples. The train split we use contains 491k images while the validation split we evaluate on contains 130k images. 

\noindent\textbf{Training:}
We train for $40000$ epochs at batch size 32 (patient count) using AdamW optimizer with a learning rate of 0.001 and a cosine decay scheduler after warmup in the first 10\% iterations. We use $n_s=n_p=n_a=2$ during training. Temperature $\tau$ is set to 0.7. Other hyperparameters follow standard settings from \cite{Jiang2023HierarchicalDL}.

\noindent\textbf{Evaluation:}
K Nearest Neighbor (kNN) classification is used to evaluate downstream performance. During evaluation, the pretrained visual backbone is frozen and representations for train and test splits are computed. The class labels of k Nearest Neighbors of the training data is used to predict the class of a given validation patch. Slide and patient level metrics are reported by average pooling the patch level prediction scores of component patches of the given slide or patient. We use all patches from OpenSRH and only 400 randomly loaded patches per slide from TCGA for evaluation. The latter limitation is due to the large slides in TCGA data.
We report kNN classification accuracy (ACC) metric per each task.

\subsection{Results Comparison}
 
We first present our results on the OpenSRH dataset in \Cref{tab:main_srh}. Our approach outperforms all prior works by a clear margin while even surpassing supervised pretraining settings for Slide Classification. 
We next evaluate on TCGA dataset and report these results in \Cref{tab:main_tcga}. Our \modelname outperforms all prior work achieving state-of-the-art results.
We take these results as indication for strong representation learning ability of our \modelname framework. 

\begin{table*}[t]
    \centering
    \caption{\textbf{Evaluation on SRH Dataset:} 
    Baseline results are reproduced on the publicly available OpenSRH dataset (a subset of the original SRH dataset) under conditions similar to \modelname. Our proposed \modelname outperforms current state-of-the-art histopathology SSL methods in all three tasks.
    }
    \label{tab:main_srh}
    \vspace{-1.0em}
    \def\arraystretch{1.0}  
    \setlength\tabcolsep{0.8em}  
    \scalebox{0.95}{
    \begin{tabular}{lccc}
    \toprule
    Method     & Patch Classification & Slide Classification  & Patient Classification  \\ \midrule
    Supervised \cite{Jiang2023HierarchicalDL}     & 88.9 & 89.0 & 93.9 \\ \midrule
    SimCLR \cite{chen2020simple}        & 81.0 & 82.1 & 87.8  \\
    SimSiam \cite{chen2021exploring}        & 80.3 & 81.4 & 86.0  \\
    BYOL \cite{grill2020bootstrap}           & 83.5 & 84.3 & 90.5  \\
    VICReg \cite{bardes2022variance}         & 82.1 & 83.4 & 87.4  \\
    HiDisc \cite{Jiang2023HierarchicalDL}        & 82.2 & 87.6 & 88.3  \\ \rowcolor{Gray}
    \modelname (ours) & \textbf{84.1} & \textbf{89.5} & \textbf{91.7} \\ \bottomrule
    \end{tabular}
    }
    \vspace{-0.5em}
\end{table*}

\begin{table*}[t]
    \centering
    \caption{\textbf{Evaluation on TCGA Dataset:} 
    Our proposed \modelname achieves state-of-the-art results across all tasks in this benchmark. 
    }
    \label{tab:main_tcga}
    \vspace{-1.0em}
    \def\arraystretch{1.0}  
    \setlength\tabcolsep{0.8em}  
    \scalebox{0.95}{
    \begin{tabular}{lccc}
    \toprule
    Method     & Patch Classification & Slide Classification  & Patient Classification  \\ \midrule
    Supervised \cite{Jiang2023HierarchicalDL}     & 85.1  & 88.3 & 88.3 \\ \midrule
    SimCLR \cite{chen2020simple}       & 77.8 & 83.0 & 80.7 \\
    SimSiam \cite{chen2021exploring}        & 68.4 & 77.2 & 76.6 \\
    BYOL \cite{grill2020bootstrap}           & 80.0 & 84.1 & 83.1 \\
    VICReg \cite{bardes2022variance}         & 75.5 & 80.8 & 77.0 \\
    HiDisc \cite{Jiang2023HierarchicalDL}        & 83.1 & 85.1 & 83.6 \\ \rowcolor{Gray}
    \modelname (ours) & \textbf{89.7} & \textbf{92.9} & \textbf{87.9} \\ \bottomrule
    \end{tabular}
    }
    \vspace{-1.5em}
\end{table*}

\subsection{Ablations}
We perform all ablations on OpenSRH dataset following the same self supervised 
pretraining stage followed by kNN evaluation as in \Cref{sec:exp}. 

\begin{table*}[t]
    \centering
    \caption{\textbf{Ablation on SSL Objectives.}}
    \label{tab:abl_SSL}
    \vspace{-1.0em}
    \def\arraystretch{1.0}  
    \setlength\tabcolsep{0.8em}  
    \scalebox{0.95}{
    \begin{tabular}{lcccc}
    \toprule
    Method & Loss   & Patch Accuracy & Slide Accuracy & Patient Accuracy \\ \midrule
    HiDisc \cite{Jiang2023HierarchicalDL} & HiDisc & 82.2  & 87.6 & 88.3 \\ 
    Ours   & HVC    & 82.6  & 89.5 & 91.7 \\ \rowcolor{Gray}
    Ours   & HLSS   & \textbf{84.1} & \textbf{89.5} & \textbf{91.7} \\ \bottomrule
    \end{tabular}
    }
    \vspace{0.0em}
\end{table*}
\begin{table*}[t]
    \centering
    \caption{\textbf{Ablation on Hierarchical Text Integration.}}
    \label{tab:abl_htext}
    \vspace{-1.0em}
    \def\arraystretch{1.0}  
    \setlength\tabcolsep{0.8em}  
    \scalebox{0.90}{
    \begin{tabular}{lcccccc}
    \toprule
    Method & Loss   & \begin{tabular}[c]{@{}l@{}}Vision \\ Hierarchy\end{tabular} & \begin{tabular}[c]{@{}c@{}}Text\\ Hierarchy\end{tabular} & \multicolumn{1}{c}{\begin{tabular}[c]{@{}c@{}}Patch \\ Accuracy\end{tabular}} & \multicolumn{1}{c}{\begin{tabular}[c]{@{}c@{}}Slide \\ Accuracy\end{tabular}} & \multicolumn{1}{c}{\begin{tabular}[c]{@{}c@{}}Patient \\ Accuracy\end{tabular}} \\ \midrule
    HiDisc \cite{Jiang2023HierarchicalDL} & HiDisc & patient  & none & 82.2 & 87.6  & 88.3 \\ \rowcolor{Gray}
    Ours   & HVC    & patient  & patient & \textbf{81.9}  & \textbf{89.1}  & \textbf{90.0} \\
    Ours   & HVC    & patient  & slide  & 81.2 & 86.5 & 86.7 \\
    Ours   & HVC    & patient & patch & 78.3 & 85.0 & 85.0 \\ \bottomrule
    \end{tabular}
    }

    \vspace{1.0em}

    \caption{\textbf{Ablation on Granular Language Descriptions.}}
    \label{tab:abl_granular}
    \vspace{-1.0em}
    \def\arraystretch{1.0}  
    \setlength\tabcolsep{0.8em}  
    \scalebox{0.95}{
    \begin{tabular}{lccccc}
    \toprule
    Method & Loss & \begin{tabular}[c]{@{}c@{}}Text\\ Hierarchy\end{tabular} & \multicolumn{1}{c}{\begin{tabular}[c]{@{}c@{}}Patch \\ Accuracy\end{tabular}} & \multicolumn{1}{c}{\begin{tabular}[c]{@{}c@{}}Slide \\ Accuracy\end{tabular}} & \multicolumn{1}{c}{\begin{tabular}[c]{@{}c@{}}Patient \\ Accuracy\end{tabular}} \\ \midrule
    Ours   & HVC  & non-granular & 81.9  & 89.1  & 90.0  \\ \rowcolor{Gray}
    Ours   & HVC  & granular & \textbf{82.6}  & \textbf{89.5} & \textbf{91.7} \\ \bottomrule
    \end{tabular}
    }
    \vspace{-1.0em}
\end{table*}

\noindent\textbf{SSL Objectives :}
In the first ablation study, we analyse the contribution of each loss component to the performance of the model. Refer table \ref{tab:abl_SSL}. \modelname with only HVC loss surpass HiDisc performace. Hierarchical projection layers initialised with granular text vectors, injecting language information to the lower dimensional spaces where hierarchical contrastive loss operates, is the contributing factor in this situation. When HVC loss is combined with a vision-text alignment loss for patch representations, it furthur improves the patch representations as observed in the results.

\noindent\textbf{Hierarchical Integration of Text :}
Integration of text hierarchically to the self-supervised setup improves performance when used at all levels as illustrated by results in table \ref{tab:abl_htext}. This study was conducted by using only the HVC loss component.

\noindent\textbf{Granular Descriptions :}
Here, we study the effect of using granular descriptions separately generated for each hierarchical level against using a generic set of textual descriptions repeated across all three levels of hierarchy. Refer table \ref{tab:abl_granular}. This study was also conducted by using only the HVC loss component and utilising all three levels in both vision and text.

\subsection{Interpretability}
Representations learnt by a language guided approach are more interpretable. We obtained a set of cancer markers specific for each tumor grading in OpenSRH from a histopathologist. These markers were not used during training. A text embedding was derived by averaging the CLIP text embeddings of 4 language descriptions generated via an LLM per each marker. Results demonstrate the close alignment between an image representation and the descriptions of the markers from the ground truth class. More details are included in \cref{app:interpret}.

\section{Conclusion}

We introduce a novel hierarchical language-tied self-supervised learning framework, named \modelname, for histopathology 
image analysis. We propose two SSL objectives, Hierarchical Vision Contrastive loss and Hierarchical Text-to-Vision Alignment loss. Our approach compliments a hierarchical vision approach by additionally exploring hierarchy in language domain. In contrast to prior vision-text alignment SSL works that require sample-specific image captions, we train our model using a set of dataset-specific text descriptions. The state-of-the-art performance of \modelname is demonstrated through evaluations on two benchmark histopathology image datasets. The results demonstrate how \modelname learns more accurate and interpretable representations. 

\section{Acknowledgment}

The computations were enabled by resources provided by the National Academic Infrastructure for Supercomputing in Sweden (NAISS) at Alvis partially funded by the Swedish Research Council through grant agreement no. 2022-06725. The results presented in this paper are in part based upon data generated by the TCGA Research Network : \href{https://www.cancer.gov/tcga}{\textcolor{blue!50!cyan}{https://www.cancer.gov/tcga}}. We thank them for their dedication to advancing cancer research. 

\newpage
%
%
\bibliographystyle{splncs04}
\bibliography{sample}

\newpage
\appendix
\section{Appendix}
\subsection{Hierarchical Image Sampling}
\label{app:hist_imaging}
\begin{table}[h]
    \centering
    {
    \setlength{\tabcolsep}{5pt}
    \begin{tabular}{c|cc}\hline
        \shortstack{Level in\\hierarchy} & \shortstack{Independant patches \\ per patient} & \shortstack{Number of positive\\ pairs per patch}\\\hline
        Patch   & $n_s \cdot n_p $ & $n_a$\\
        Slide   & $n_s$ & $ n_p \cdot n_a$\\
        Patient & $1$ & $n_s\cdot n_p \cdot n_a$\\\hline
    \end{tabular}
    }
    \vspace{0.5em}
    \caption{\textbf{Positive pairing for hierarchical contrastive objective} $n_s$, number of slides sampled per patient, $n_p$, number of patches sampled per slide, $n_a$, number of augmentations performed on each patch.}\label{tab:pp}
\end{table}

\subsection{Interpretability}
\label{app:interpret}
\begin{figure}[ht]
    \centering
    \includegraphics[width=0.95\linewidth]{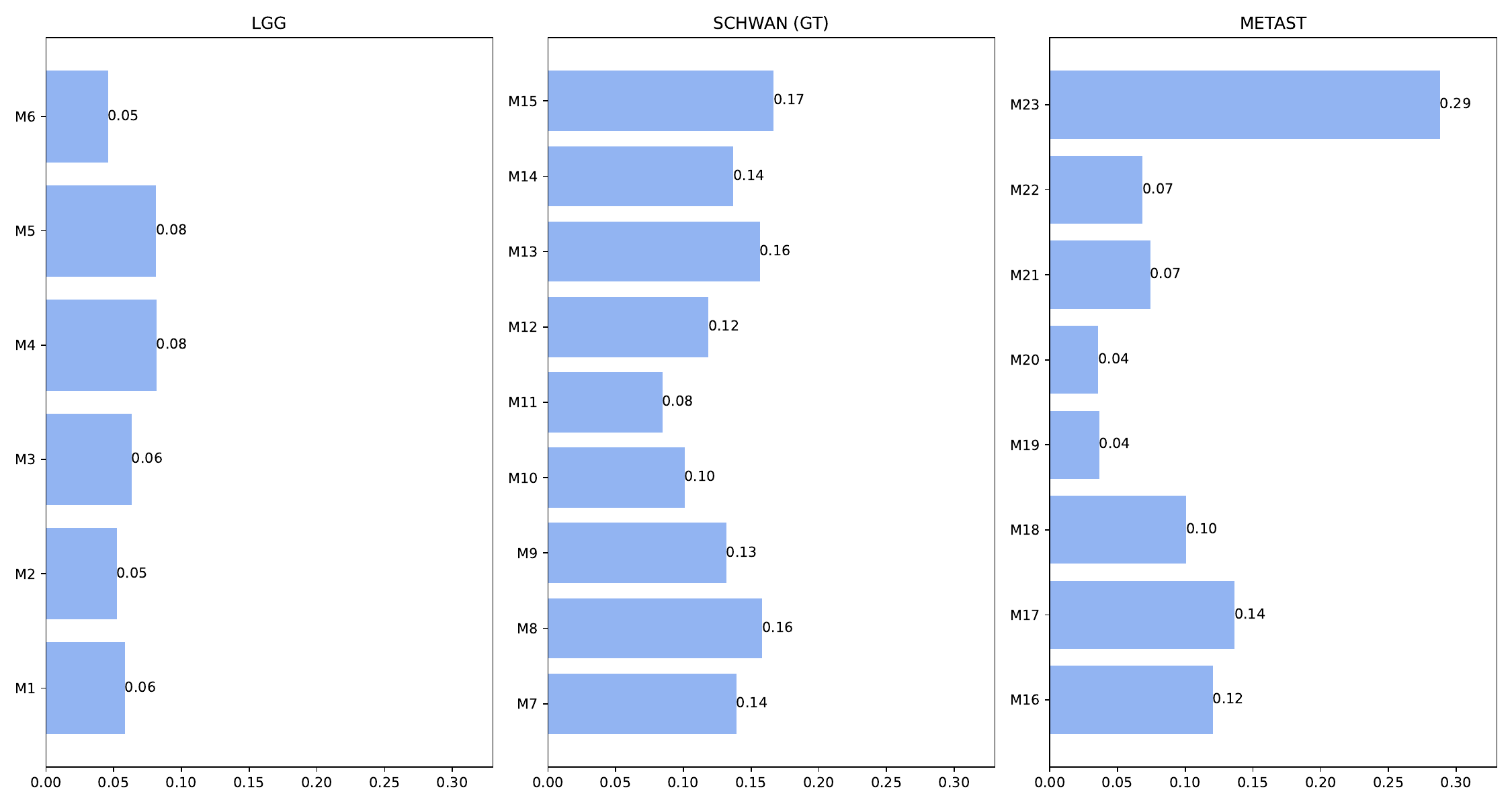}
    \vspace{-1.0em}
    \caption{
    Cosine similarity of a given sample representation with class specific cancer markers ($M_{i}$). Markers of the ground truth class (GT) are relatively more similar.
    }
    \label{fig:interpret1}
    \vspace{-1.5em}
\end{figure}

\begin{figure}[t]
    \centering
    \includegraphics[width=0.95\linewidth]{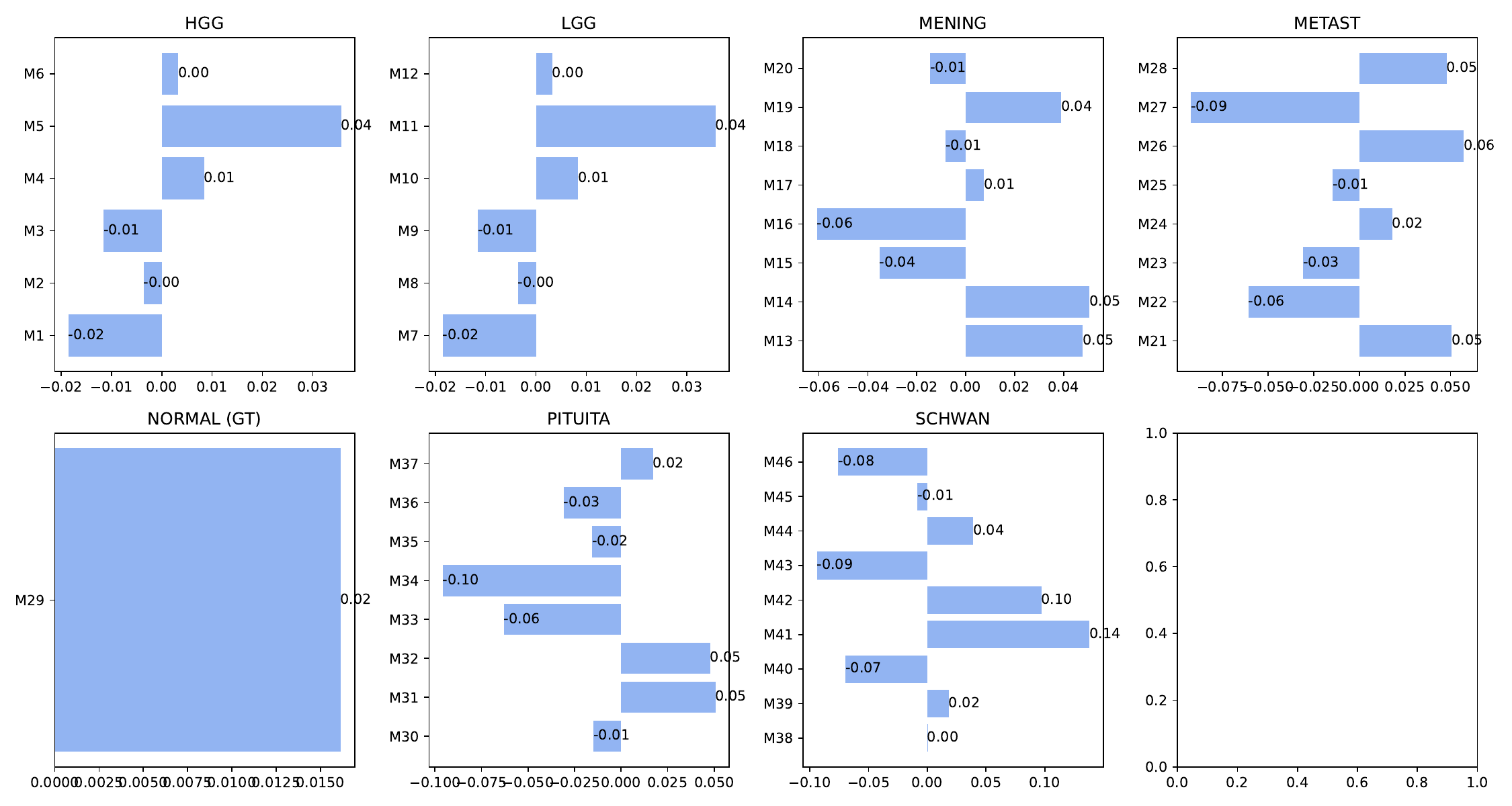}
    \vspace{-1.0em}
    \caption{
    Cosine similarity of a `normal' tissue sample representation with class specific cancer markers ($M_{i}$) of all classes is illustrated. Observe how the representation doesn't align well with majority markers in any class.
    }
    \label{fig:interpret2}
    \vspace{-1.5em}
\end{figure}

\begin{figure}[ht]
    \centering
    \includegraphics[width=0.95\linewidth]{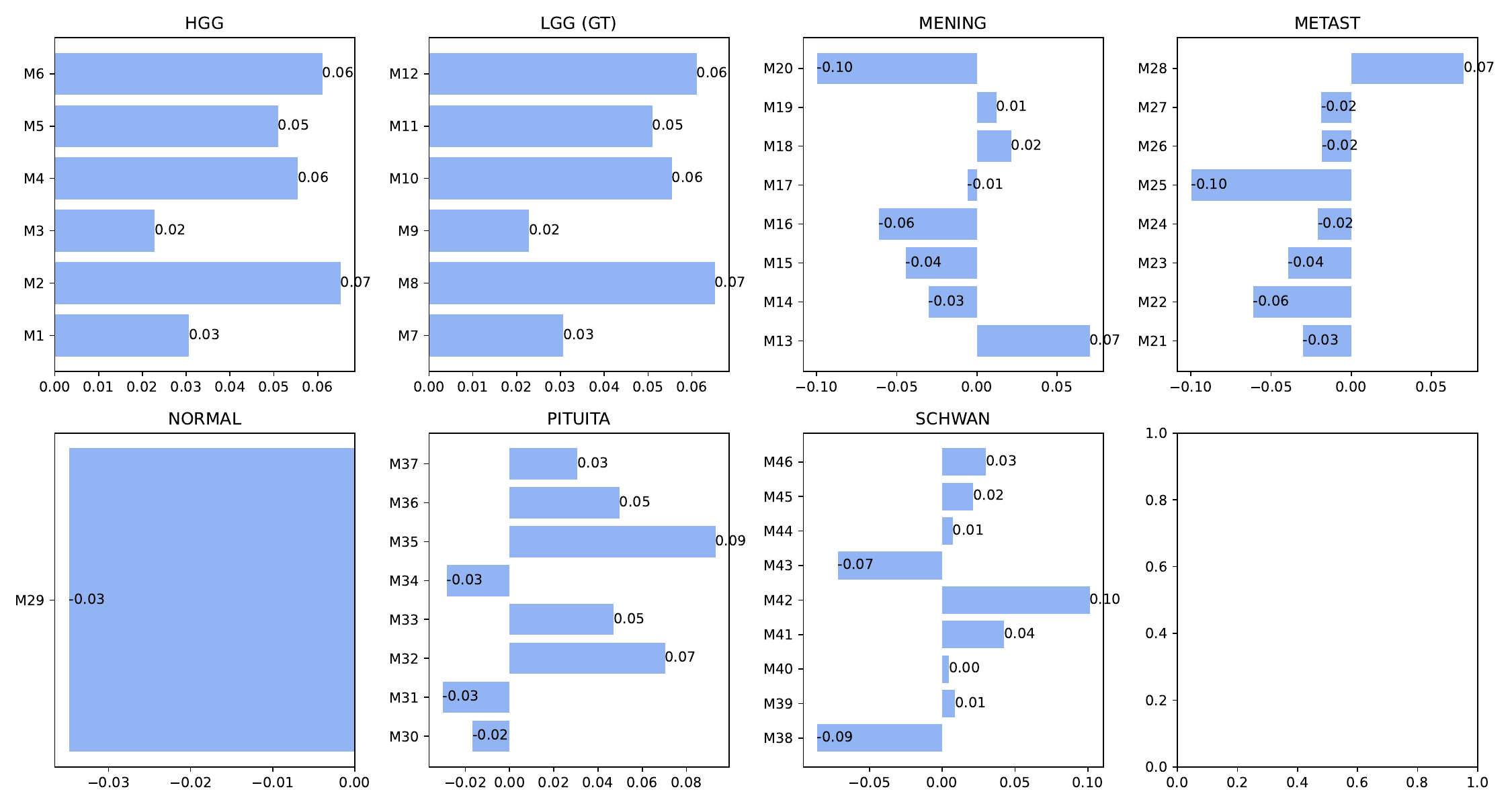}
    \vspace{-1.0em}
    \caption{
    Cosine similarity of a LGG (lower grade glioma) sample representation with class specific cancer markers ($M_{i}$). A proper alignment can only be seen in HGG \& LGG classes as they both consider similar characteristics as per the markers we obtained from an expert histopathologist.
    }
    \label{fig:interpret3}
    \vspace{-1.5em}
\end{figure}

\end{document}